\documentclass[]{IEEEtran}

\usepackage{graphicx}    
\usepackage{url}
\usepackage{color}
\usepackage{subcaption}
\usepackage{todo}

\makeindex

\begin{document}

\title{View Planning and Navigation Algorithms for Autonomous Bridge Inspection with UAVs}

\author{Kevin Yu, Prajwal Shanthakumar, Jonah Orevillo, Eric Bianchi, Matthew Hebdon, Pratap Tokekar%
\thanks{K. Yu, P. Shanthakumar, J. Orevillo, Eric Bianchi, Matthew Hebdon, and P. Tokekar were with the Department of Electrical and Computer Engineering, Virginia Tech, Blacksburg, VA 24061 USA when this work was performed. E-mail: klyu@vt.edu, prajwal@vt.edu , j0n4h0@vt.edu, beric7@vt.edu , mhebdon@vt.edu, \& tokekar@vt.edu.}}

\maketitle

\begin{abstract}
We study the problem of infrastructure inspection using an Unmanned Aerial Vehicle (UAV) in box girder bridge environments. We consider a scenario where the UAV needs to fully inspect box girder bridges and localize along the bridge surface when standard methods like GPS and optical flow are denied. Our method for overcoming the difficulties of box girder bridges consist of creating local navigation routines, a supervisor, and a planner. The local navigation routines use two 2D \textsc{Lidar}s for girder and column flight. For switching between local navigation routines we implement a supervisor which dictates when the UAV is able to switch between local navigation routines. Lastly, we implement a planner to calculate the path along that box girder bridge that will minimize the flight time of the UAV. With local navigation routines, a supervisor, and a planner we construct a system that can fully and autonomously inspect box girder bridges when standard methods are unavailable.
\end{abstract}

\IEEEpeerreviewmaketitle

\section{Introduction}
Highway bridges are typically required to be inspected every two years~\cite{FederalRegister}. 
Current methods of bridge inspection takes significant effort from human operators who have to go through extensive training to become certified. These human operators may have to don harnesses and operate cranes to fully inspect a bridge. This often requires closure of roads and expensive equipment, and often places human inspectors in potentially dangerous situations. The complexity and the sheer number of bridges, coupled with the requirement of highly-trained inspectors and hefty insurance can pose significant challenges for cost-constrained owners. 
A more promising alternative is robots that can carry out routine inspection without affecting the flow of traffic or posing a risk to human inspectors that often have to be suspended at dangerous heights. With robotic technology maturing and commercial solutions entering the market, Unmanned Aerial Vehicles (UAVs) can make bridge inspection faster, safer and less expensive~\cite{zink2015unmanned}.

Even with the development of UAV technology, manual piloting UAVs can still be difficult and requires specialized training. Challenges include flying without line-of-sight, especially along bridges and under the deck; hovering in place for long periods of time in windy conditions; and operating without GPS and compass measurements~\cite{gillins2016cost}. The GPS reception around bridges is typically noisy, if not completely absent. Furthermore, compass can be unreliable since bridges have significant metal in its structure. Instead, we need autonomous navigation and control algorithms that can stability the UAVs in high wind conditions that do not rely on GPS or compass information.

Ultimately, the goal is to plan paths of the UAV so as to obtain high-quality images of specific points of interest. Even if the local navigation problem is solved, waypoint planning in the global frame of reference is a challenge. Choosing waypoints can be tedious and requires precise geometric models of the bridge within some global frame. As an example, altitude profiles of bridges are not constant since many bridges slope upwards or downwards and the terrain around the bridges may be uneven. Bridges are often over water which may impede a UAV's ability to accurately measure its relative altitude. Therefore, we need algorithms that can plan inspection paths for the UAV without too much a priori global knowledge about the structure of the bridge. 

The goal of our work is to design autonomous planning and navigation algorithms to make bridge inspection with UAVs easier than the status quo. We focus primarily on box girder bridges (Figure~\ref{nav_routines}). The main contributions of this paper are a set of local navigation routines, a supervisor for switching between local navigation routines, and a planner to find the optimal sequence of local routines to inspect the bridge. We create methods that allow for flight relative to the bridge structure as well as in GPS and compass denied settings.

Similar works study infrastructure inspection~\cite{sa2014vertical, subhan2014study, yoder2016autonomous, ozaslan2017autonomous, alexis2016aerial, hollinger2017active, metni2007uav, oh2009bridge}. Sa and Corke~\cite{sa2014vertical} inspect vertical pole-top structures such as light posts. Subhan and Bhide~\cite{subhan2014study} present work that inspects underground coal mines using autonomous teams of robots. Scherer and Yoder~\cite{yoder2016autonomous} conducts autonomous flight for building a 3D model of arbitrary structures outdoors. {\"O}zaslan et al.~\cite{ozaslan2017autonomous} presents a method for autonomous navigation of penstocks and tunnels. The authors present a new approach for state estimation, mapping and shared control with Micro Aerial Vehicles (MAVs). Alexis et al.~\cite{alexis2016aerial} present a unique approach to infrastructure inspection by using contact-based inspection as the driving motivation to the paper. Hollinger et al.~\cite{hollinger2017active} presents a method for underwater inspection by actively choosing viewpoints of an object to increase performance of inspection. Metni and Hamel~\cite{metni2007uav} look at bridge inspection using UAVs. The work focuses on UAV flight along a planar target using a control law that is based on computer vision. Oh et al.~\cite{oh2009bridge} looks at bridge inspection using a vision sensor attached on the end of a snooper truck. By doing this the authors mitigate all risks to the human operator by replacing them with robotic sensors.

All of these works implement methods for infrastructure inspection, but some lack high-level planning that our paper looks to address. The works of~\cite{sa2014vertical, subhan2014study, ozaslan2017autonomous, metni2007uav,oh2009bridge} all do infrastructure inspection, but do not implement a high level planner. Sa and Corke~\cite{sa2014vertical} implement controls that help to assist the UAV operator, while Subhan and Bhide~\cite{subhan2014study} propose an architecture that can be used for search and rescue robots. {\"O}zaslan et al.~\cite{ozaslan2017autonomous} does not need a high level planner because they fly along the inside of a single tunnel. Metni and Hamel~\cite{metni2007uav} and Oh et al.~\cite{oh2009bridge} both conduct bridge inspection, but only look at implementing control laws and implementing better imaging for crack detection, respectively.

A few of the mentioned works look at planning paths for the UAV for infrastructure inspection. Scherer and Yoder~\cite{yoder2016autonomous} plan paths incrementally along arbitrary structures. This however may lead to a non-optimal path of the arbitrary structure. We are able to guarantee an optimal solution for coverage of a 3D surface. Alexis et al.~\cite{alexis2016aerial} plan by combining Traveling Sales Person (TSP) and Rapidly-exploring Random Tree (RRT*) to obtain a close-to-optimal solution. Our algorithm guarantees optimality as well as conducts path planning for UAVs that are not in contact with the structure of interest. Hollinger et al.~\cite{hollinger2017active} implement two planners for non-adaptive and adaptive classification. The authors use object detection to decide on view points that would best help to classify objects or defects on a structure. This planner looks at planning for a singular area to get the best view where as our planner looks at optimal planning for a large 3D structure like a bridge.

These works all illustrate inspection of different structures as well as some of the difficulties that arise when conduction infrastructure inspection. When conducting infrastructure inspection many standard methods for autonomous movement do not work such as optical flow or GPS due to the featureless surfaces that are needed for optical flow and  unreliable GPS close to structures. All the works exemplify reasons why optical flow will be difficult. When inspecting structures at a close distance such as a light post or walls in contact inspection, the features are minimal or even non-existence for visual navigation methods like optical flow. In works~\cite{subhan2014study, ozaslan2017autonomous, hollinger2017active} it would be impossible to conduct UAV autonomous flight with GPS. Many environments make it impossible for conducting autonomous flight using GPS, but there are also environments that lead to unreliable GPS such as under bridge surfaces.

Our work implements a low-level planner as well as a high-level planner. Our low-level planner look at using \textsc{Lidars} and our high-level planner looks at using a Generalized Traveling Sales Person (GTSP) approach. Earlier works have shown that solving the GTSP instance of a problem can provide solutions that are comparable if not the same as the optimal solution that a TSP solver may provide while greatly reducing the time taken to solve~\cite{Smith2016GLNS, mathew2015multirobot, tokekar2016visibility}. Noon and Bean~\cite{noon1993efficient} provide a method for reducing the GTSP instance into a TSP instance, allowing us to solve the problem to optimality. For our work we can guarantee optimality by executing the conversion characterized by Noon and Bean as well as use GTSP solvers like Generalized Large Neighborhood Search (GLNS) to obtain close-to-optimal solutions in reasonable amounts of time.

This paper expands the preliminary version presented at ISER~\cite{shanthakumar2018view} and includes new full-scale simulations and complete implementation of GTSP planner.

The rest of the paper is organized as follows. Section \ref{tech} describes the hardware and the software framework of the proposed system. Section \ref{exp} presents experimental and simulation results. We conclude in Section \ref{future} with a discussion of the challenges and future work.

\section{System Description}\label{tech}
In this section, we describe the hardware and the software architecture of the proposed system.

\subsection{Hardware Description}\label{system}
\begin{figure}[hbt!]
\centering
\includegraphics[width=\columnwidth]{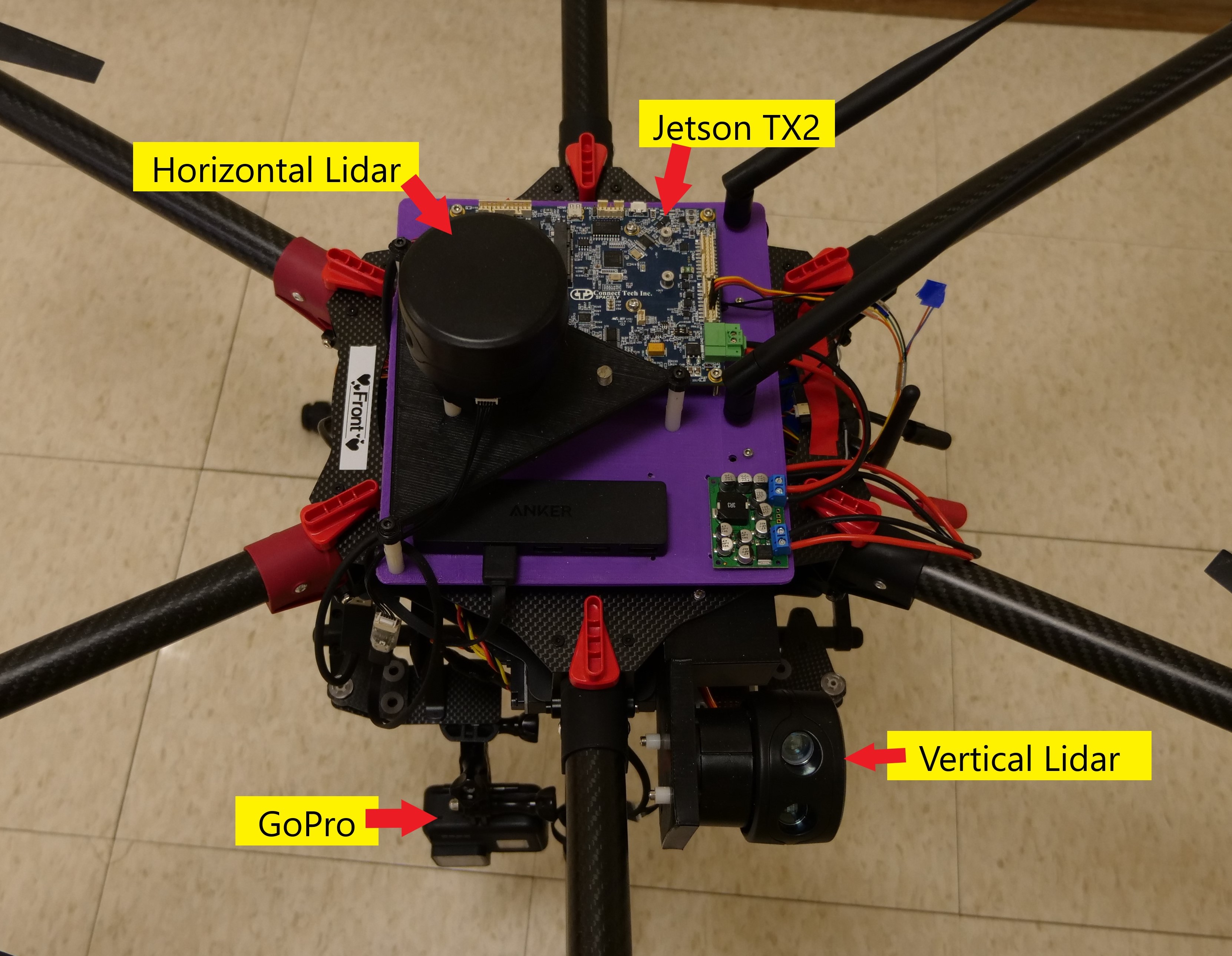}
\caption{UAV for bridge inspection.}
\label{quad}
\end{figure}

We use the DJI S900~\cite{Spreadin77:online} as our platform (Fig.~\ref{quad}). This UAV has a hexarotor design. Six motors provide a maximum thrust of 2.5Kg. The UAV is powered by a 6S 10000mAh Lithium-polymer battery and weighs 3.5kg without any payload. 

We use the Pixhawk autopilot as the main flight controller~\cite{CubePixh75:online}. The Pixhawk autopilot has an in-built IMU, compass, accelerometers, and gyroscopes. The Pixhawk runs the PX4 firmware version 1.7.3v~\cite{meier2015px4} for low-level flight control. The UAV is also equipped with an \textsc{Nvidia} Jetson TX2 for high-level control. The software on the Jetson runs on Ubuntu 16.04 and uses  \textsc{ROS} (Robot Operating System) Kinetic 1.12.13. The \textsc{Nvidia} Jetson TX2 is used for processing sensor data and publishing the desired velocity to the Pixhawk flight controller. The \textsc{Nvidia} Jetson TX2 has a 256-core Pascal GPU, quad-core ARMv8 processor rev3, 8GB of DDR4 memory, 32GB of memory, and uses a maximum of 15W of power. 

The UAV is equipped with two Scanse Sweep 2D \textsc{Lidar} sensors~\cite{scanse} and a GoPro Hero7 Black Camera. The two \textsc{Lidar}s are placed in a horizontal and vertical arrangement. Both \textsc{Lidar}s are used for estimating the distance to the bridge structure and enable autonomous navigation. These sensors have a maximum range of 40m, maximum resolution of 1cm, sample rate of up to 10Hz, $360^{\circ}$ field of view, use 5V@650mA, and weigh 120g each. GoPro Hero7 Black Camera is used to collect images of the bridge structure for defect identification. The GoPro Hero7 Black shoots 4K 60FPS video with an aspect ration of 16:9. This camera can be replaced by a higher resolution inspection-grade camera.

We use the Pixhawk to run PX4 firmware along with the mavros package~\cite{mavrosRO76:online}. Mavros is a ROS package that allows for communication between the PX4 firmware and the on board \textsc{Nvidia} Jetson TX2. The Jetson TX2 runs software associated with the local navigation routines and the higher-level supervisor. The software is modular with separate ROS nodes running in \textsc{ROS} Kinetic. The individual modules are explained in Section~\ref{approach}. ROS nodes allow for the publishing and subscribing of data. The supervisor and local navigation routines communicate with each other through publishing and subscribing to figure out what velocity commands need to sent to the flight controller. Once the correct velocity is figured out the give velocity is broadcast to the UAV flight controller through mavros. 

\subsection{Algorithm Description}\label{approach}
Our current approach for complete inspection coverage of all bridge surfaces is to autonomously execute a series of maneuvers from a library of navigation routines (Fig. \ref{nav_routines}). While the UAV autonomously navigates the bridge, an onboard camera records images of the bridge that can be examined for defects in real time and/or used for post-processing.

The algorithm consists of three modules: 
\begin{enumerate}
    \item global planner to find the sequence of local navigation routines needed for complete visual coverage of the bridge;
    \item local navigation routines for real-time, low-level, \textsc{Lidar} based navigation; and
    \item supervisor to determine if we have completed a local navigation routine and can progress to the next one using \textsc{Lidar} data.
\end{enumerate}   
We describe each of the three modules in details next. 

\subsubsection{Global Planner} \label{sec:planner}
\begin{figure}[hbt!]
\centering
\begin{subfigure}[b]{0.45\textwidth}
\centering
\includegraphics[width=\columnwidth]{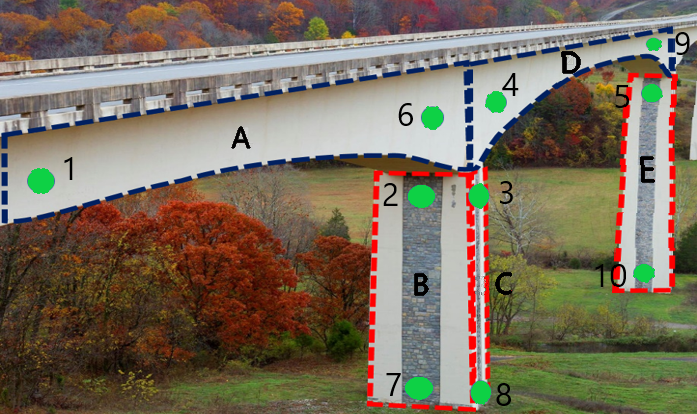} 
\caption{}
\label{sampleBridge}
\end{subfigure}
\hfill
\begin{subfigure}[b]{0.45\textwidth}
\centering
\includegraphics[width=\columnwidth]{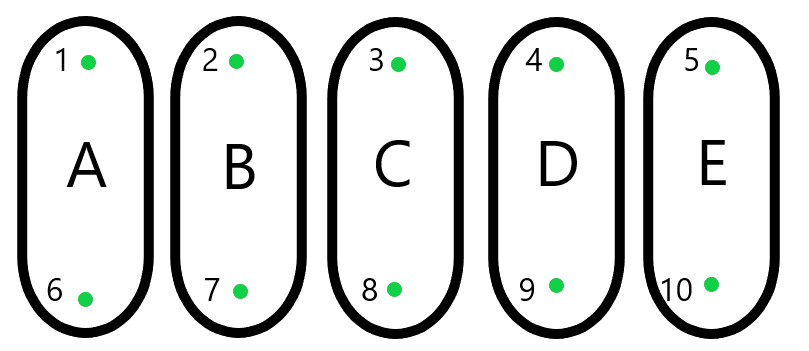} 
\caption{}
\label{GTSP}
\end{subfigure}
\caption{(a) Bridge partitioned into surfaces with entry/exit nodes (green dots). (b) GTSP clusters with entry/exit nodes (green dots) corresponding to Fig.~\ref{sampleBridge}.}
\end{figure}

The first step in our algorithm is to find a global tour that the UAV must follow for full visual coverage of the bridge. In this work, we focus on inspecting box girder bridges (shown in Figure~\ref{sampleBridge}), in particular all the external surfaces of the bridge (girder, column, deck, etc.). 

We first partition the bridge surfaces into a set of planar surfaces, as shown in Fig.~\ref{sampleBridge}. Each planar surface can be approximated by a polygon. We can fully inspect the bridge by finding a route that inspects each of the surfaces. This decomposition makes it easier to find a global plan as well as make it easier for the operator to understand the plan. 

To inspect each surface, we need to visually cover the corresponding polygon. The local navigation controller ensures that the robot moves in such a way so as to visually cover each polygon. Therefore, the goal of the planner is to find the sequence in which to visit the polygons as well as the entry and exit points for each polygon. 
Here, we take advantage of the specific structure of the box girder bridge. We associate with each polygon two points, which will represent coverage of the polygon from both directions. This means that one of the nodes will be chosen as the entry node to the planar surface and the other node is chosen as the exit node. The local navigation routine controller will ensure navigation between the entry and exit nodes and ensure visual coverage of the bridge surface. Since there are no physical markings on the bridge surface, the supervisor will identify when the exit node of the current polygon/entry node of the next polygon has been reached and pick the appropriate local navigation routine to continue rest of the visual inspection tour.

The visual coverage problem is that of finding the sequence in which the bridge surfaces must be traversed, as well as determining the entry node for each of the polygons representing the surfaces. We formulate visual coverage as a Generalized Traveling Salesperson Problem (GTSP)~\cite{noon1993efficient}. The input to GTSP is a graph where the nodes are partitioned into disjoint clusters. The goal is to find the minimum cost tour that visits at least one node in each cluster. GTSP generalizes the NP-Hard Traveling Salesperson Problem, and therefore is NP-Hard as well. Nevertheless, there are good numerical solvers present~\cite{Smith2016GLNS} that can be used to find the optimal solution for reasonably-sized instances. 

We convert the visual coverage problem into a GTSP instance as follows. We create one cluster for each of the polygons. Each cluster contains the two nodes associated with that polygon, as shown in Fig.~\ref{GTSP}. One of the two nodes will be chosen as the entry node and the other will be chosen as an exit node by the algorithm. An edge is created between every pair of nodes that belong to separate clusters. These edges are the combination of coverage of the cluster and then traveling to another cluster. In Fig.~\ref{GTSP} if we were to go from node 1 to node 7 we would have to cover cluster A and then go to node 7 by going from node 1 to node 6 then to node 7. Once that is done the algorithm filters out edges that can not be traversed by one of the local navigation routines.

For any cluster X, let $X_1$ and $X_2$ represent its entry and exit nodes respectively. Then, the cost on an edge from node $A_1$ in cluster A to node $B_1$ in cluster B is given by, $$ C(A_1,B_1) = D(A_1, A_2) + D(A_2, B_1)$$ where $D(A_2,B_1)$ represents the distance between nodes $A_2$ and $B_1$ in 3D space, i.e., the cost is the sum of the distance to cover the current polygon (surface) and the distance to reach the entry node of the next polygon from the exit node of the current polygon. When conducting experiments a picture of the bridge is used to obtain an estimate of distances. Using rough estimates of the distances between entrance and exit nodes for each planar surface will still allow us to obtain an accurate solution as long as it is to scale.

Instead of using the actual 3D distances (which will not be known when we do not have the 3D model of the bridge), we can use estimates of the distances. In this work, we use a side-view image of the bridge to find the scaled distances (in pixel units) between the nodes. 

This is the input to the GTSP solver. We use the Generalized Large Neighborhood Search (GLNS) solver~\cite{Smith2016GLNS} which is the state-of-the-art for GTSP instances. Once we obtain a solution from the GLNS solver we can obtain the full tour to be visited by the UAV. The GLNS solver will give us a solution in the GTSP instance of our problem. To convert that back to our original problem we look at the order in which the clusters are visited and the entry points, which both are given by the solver. In Fig.~\ref{GTSP} if the output from the GLNS solver was 1, 2, 8, 4, 5 then we know that we are visiting the clusters in the order of A, B, C, D, E. However we need to transform the output from the GLNS solver into something that the UAV can use for navigation along the bridge surface. To do this we take each point in the output from GLNS and add the corresponding exit point. For our output we would change the output to \textbf{1}, 6, \textbf{2}, 7, \textbf{8}, 3, \textbf{4}, 9, \textbf{5}, 10, with the bolded values as the original GLNS output. Now that we have the output we can give it to the UAV for full traversal of the bridge. By using a GTSP formulation, we can guarantee that we visit every cluster exactly once, which implies we visit each polygon surface that needs to be inspected exactly once. The local navigation routine ensures that each surface is inspected as required.

\subsubsection{Local Navigation Routines}\label{lidar}
\begin{figure}[hbt!]
\centering
\includegraphics[width=\columnwidth]{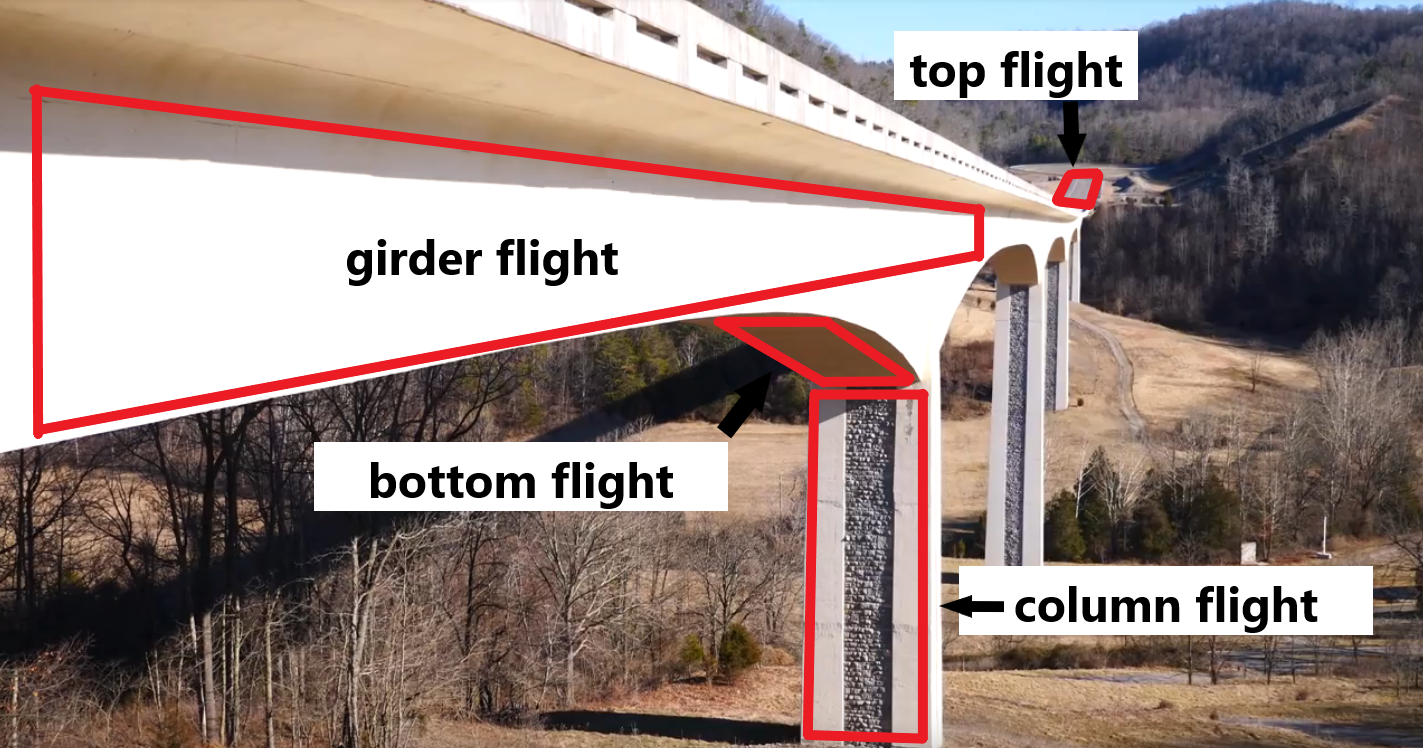}
\caption{Bridge showing the different possible local navigation routines. For girder flight it extends along the full bridge and is replicated for the opposite side. Bottom flight is along all bottom portions of the bridge as well as top flight is along the full top of the bridge. Column flight is also duplicated along all column surfaces.}
\label{fig:LNR}
\end{figure}

For safe and accurate autonomous traversal of bridge surfaces, we employ an algorithm that must be fast enough to run in real-time as well as robust to factors such as wind disturbances, unreliable GPS, and inaccurate compass measurements. When examining girder style bridges there are four main parts that need to be traversed the girder, column, bottom, and top, shown in Fig.~\ref{fig:LNR}. To traverse all of these parts of the bridge there is a total of eight local navigation routines. To traverse the girder of a bridge there are two methods that need to be employed and for traversal of the column there are two. Also when inspecting the top and bottom of the bridge surface there are two methods for each of them. For girder flight, Fig.~\ref{fig:LNR}, a right and left local navigation routine and for column flight, Fig.~\ref{fig:LNR}, up and down local navigation need to be executed. Similar to the girder local navigation routines the top flight and bottom flight, Fig.~\ref{fig:LNR}, of the bridge needs right and left local navigation routines. In total that leaves us with girder right(\textit{GR}), girder left(\textit{GL}), column up(\textit{CU}), column down(\textit{CD}), bottom right(\textit{BR}), bottom left(\textit{BL}), top right(\textit{TR}), and top left(\textit{TL}). For this paper we focus on coverage of one side face of the bridge. Through the rest of the paper we will be using the above notation. Due to this we only use four local navigation routines (\textit{GR}, \textit{GL}, \textit{CU}, and \textit{CD}).

\begin{figure}
\centering
\begin{subfigure}[b]{0.45\textwidth}
\centering
\includegraphics[width=\columnwidth]{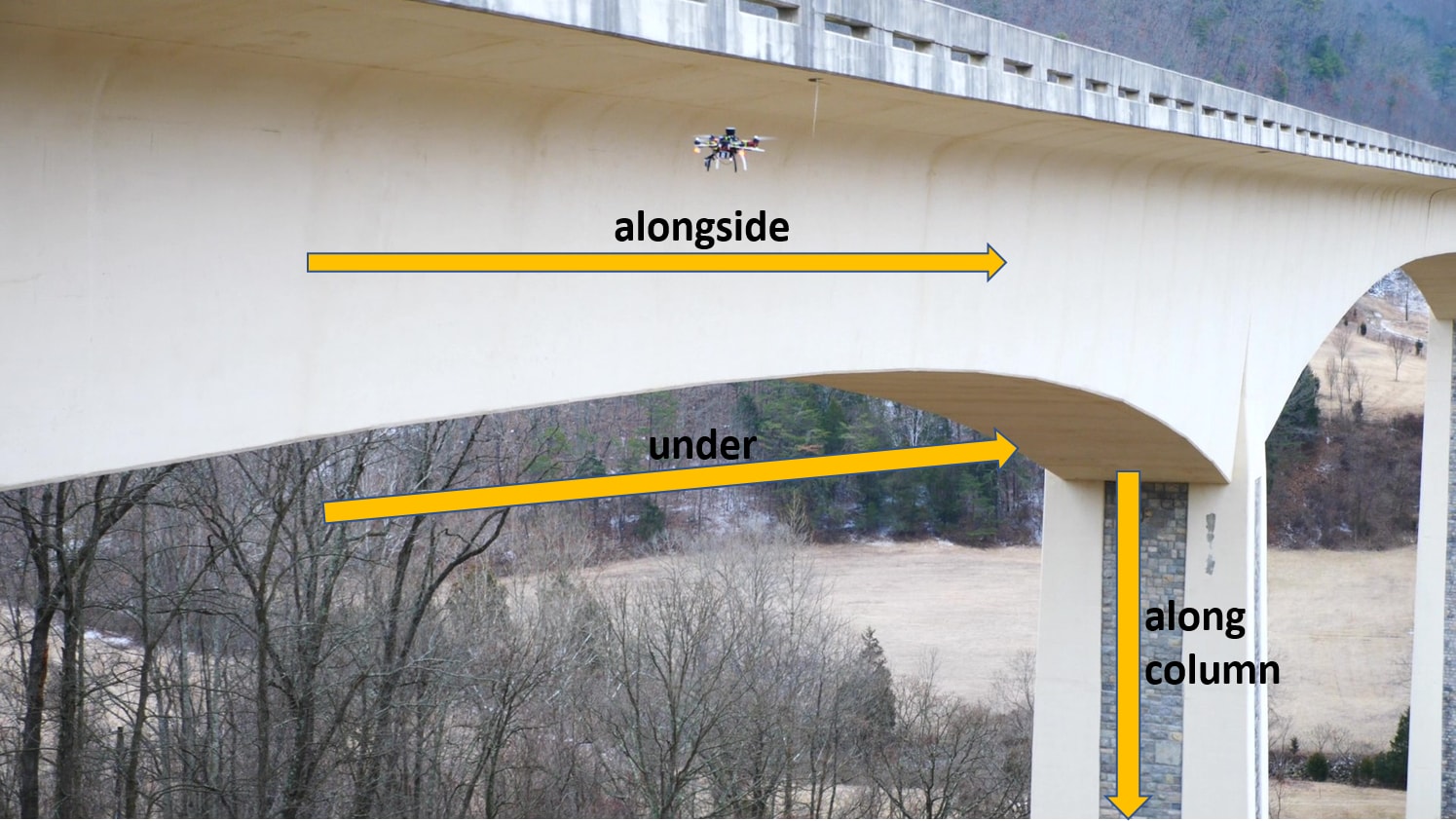} 
\caption{}
\label{nav_routines}
\end{subfigure}
\hfill
\begin{subfigure}[b]{0.45\textwidth}
\centering
\includegraphics[width=\columnwidth]{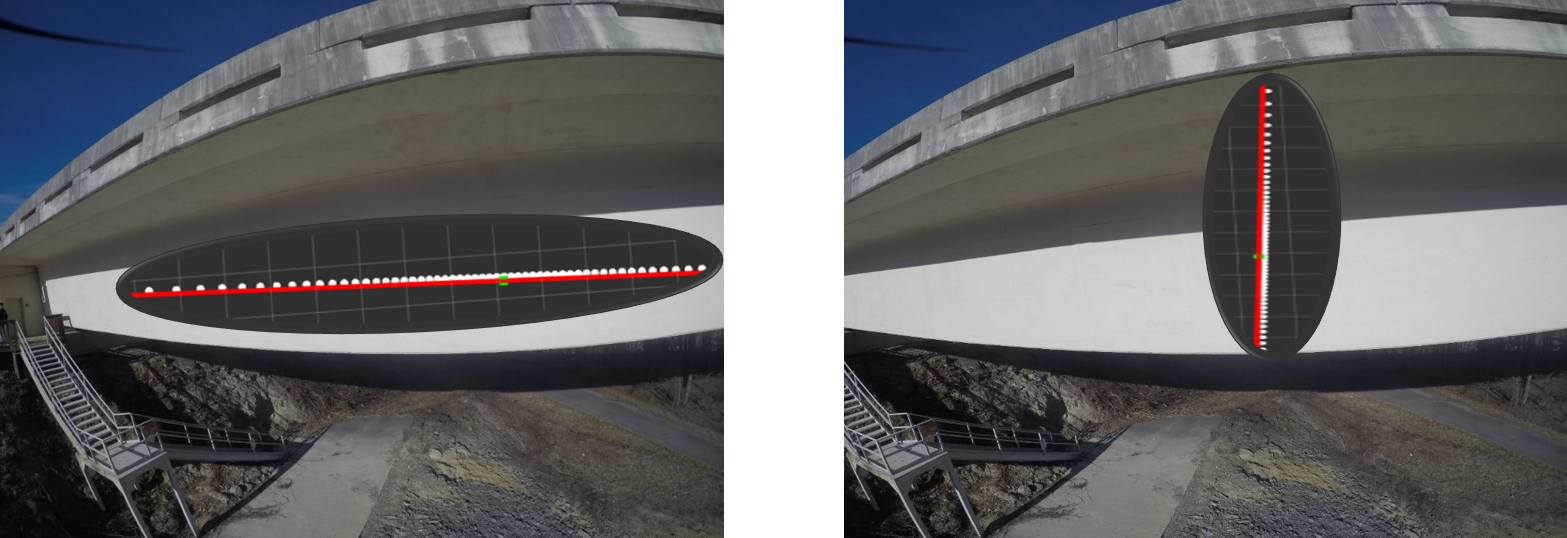}
\caption{}
\label{hscan}
\end{subfigure}
\caption{(a) Illustration of some useful navigation routines. (b) White dots: \textsc{Lidar} data; Red line: best fit using Hough Transform.}
\end{figure}

We employ 2D \textsc{Lidar} (a rotating 1D \textsc{Lidar}) based local control for this purpose. By using two 2D \textsc{Lidar}s, one placed horizontally and one placed vertically, we can obtain real-time data as well as have robust navigation. We implement all routines for traversing along the girder of the bridge and along the columns of the bridge. The four local navigation routines only use the two 2D \textsc{Lidar}s, allowing the UAV to navigate in environments that contain no GPS signal. We exploit the geometry inherent to bridges and treat the bridge surfaces as planes. A 2D \textsc{Lidar} scan of a plane shows up as a line as shown in Fig.~\ref{hscan}.

\begin{figure}[hbt!]
\noindent
\centering
\includegraphics[width=0.9\columnwidth]{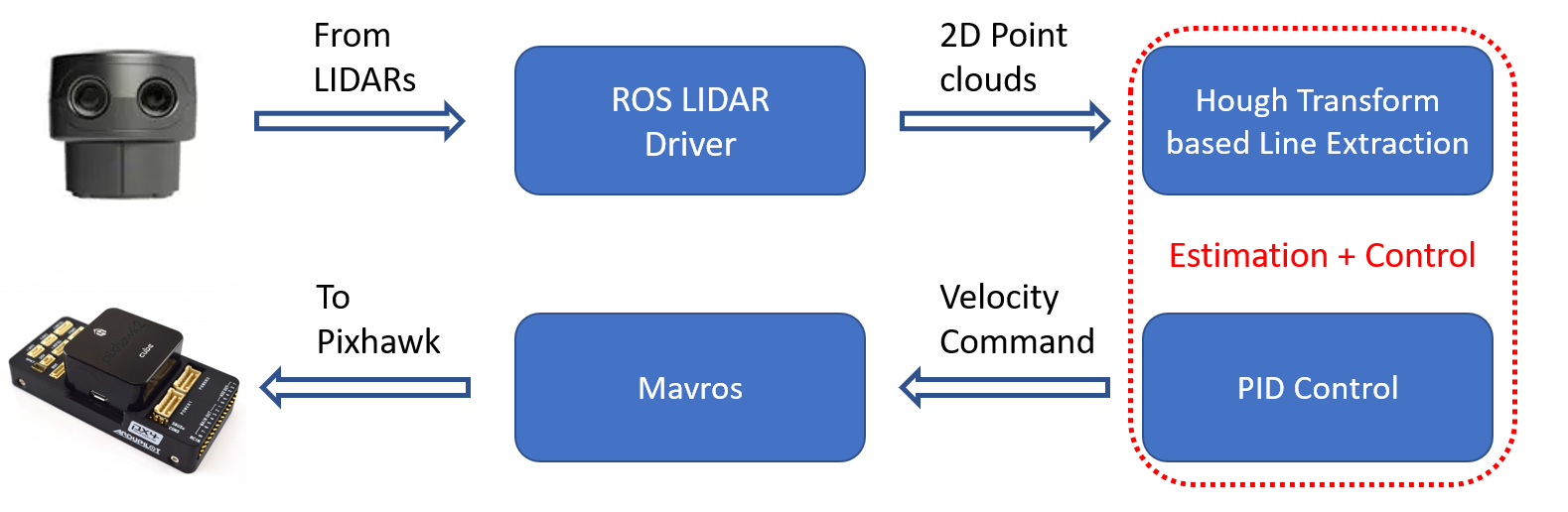} 
\caption{Software architecture for autonomous navigation.}
\label{arch}
\end{figure}

A Hough Transform based approach is used to find the best fit line that approximates a bridge surface. We implement a PID control to fly the UAV parallel to the bridge surface, maintaining a fixed distance as well as a specific distances from the top and side of the girder and column respectively. The software architecture is illustrated in Fig.~\ref{arch}. A more detailed description of Estimation and Control is provided below.

\subsubsection*{Estimation}
\begin{figure}[hbt!]
\centering
\begin{subfigure}[b]{0.35\textwidth}
\centering
\includegraphics[width=\columnwidth]{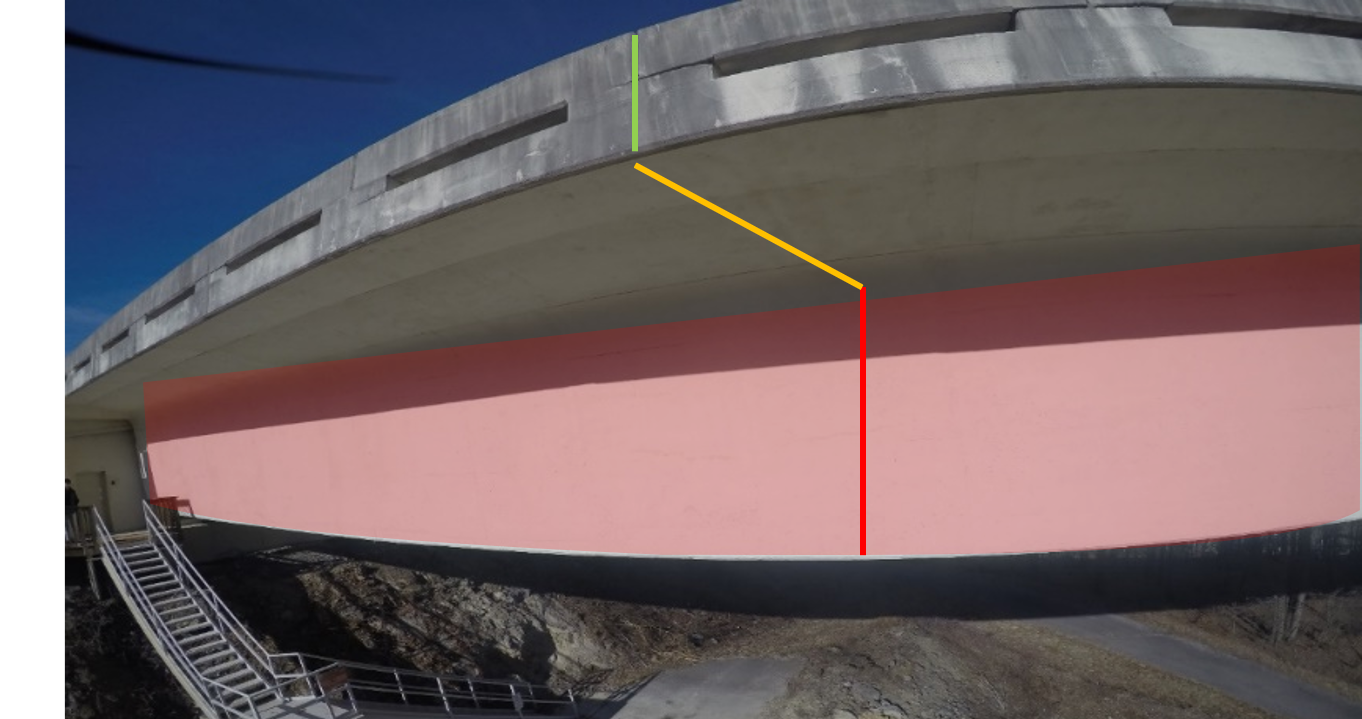}
\caption{}
\label{girderS}
\end{subfigure}
\hfill 
\begin{subfigure}[b]{0.25\textwidth}
\centering
\includegraphics[width=\columnwidth]{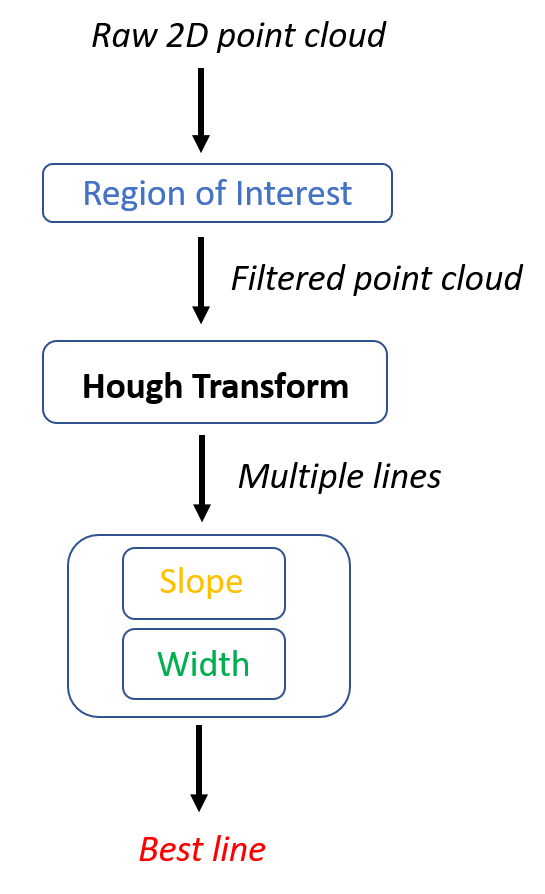}
\caption{}
\label{flow}
\end{subfigure}
\hfill
\begin{subfigure}[b]{0.35\textwidth}
\centering
\includegraphics[width=\columnwidth]{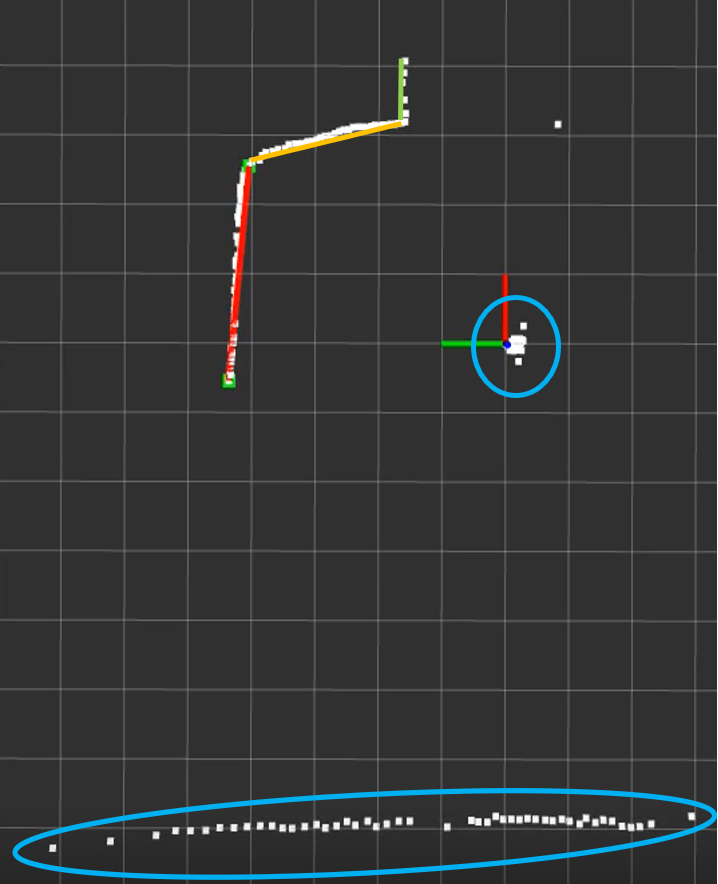} 
\caption{}
\label{filtering}
\end{subfigure}
\caption{(a) The bridge girder surface highlighted in red. (b) Steps involved in the Hough Transform based Line Extraction process. (c) Visualization (rviz) of filtered out points and extracted lines.}
\end{figure}

To understand how bridge surfaces are estimated, consider, the problem of estimating the vertical span of the bridge girder surface (highlighted in red in Fig.~\ref{girderS}). We start by obtaining the 2D point cloud data from the vertical \textsc{Lidar}. Once we get the data we start to estimate the vertical span of the bridge girder surface. The following are the steps involved (explained with reference to Figs.~\ref{girderS}, ~\ref{flow} \& \ref{filtering}). Data points that are very close to the body of the UAV and very far from the body of the UAV (ground plane, distant trees, etc.) are initially filtered out (circled in blue in Fig.~\ref{filtering}). Then we execute a Hough Transform to extract lines from the filtered data. Let us assume the three lines as shown in Figs.~\ref{girderS} \& \ref{filtering} are found with enough confidence. The confidence is directly related to the number of raw data points that fall on the extracted lines. We are interested in extracting only the red line which approximates the vertical span of the bridge girder surface. We expect this line to have the characteristics of $\approx{90^\circ}$ and width $3-5$m. Due to this, the yellow line with slope $\approx{10^\circ}$ and the green line with width $\approx{1}$m will be filtered out because the characteristics do not match up with the expected vertical span of the bridge (Figs.~\ref{girderS} \& \ref{filtering}). These expectations are specific based on what bridge is being traversed and will need to be changed based on the current bridge. However we do not need to know these accurately and only need an estimate to allow for the UAV to navigate the bridge. 

\subsubsection*{Control}
The goal of our algorithm is for the UAV to fly parallel to the (estimated) bridge surfaces. This involves implementing two independent PID loops along two perpendicular axes to maintain position with respect to the bridge surfaces. For \textit{GR} and \textit{GL} flight, the UAV maintains altitude and distance relative to the bridge. For \textit{CU} and \textit{CD}, flight the UAV maintains distance from the column edge and distance relative to the bridge. Dependent on the desired local navigation routine a constant nominal velocity along the axis of movement maneuvers the UAV parallel to the bridge surface, right and left for girder flight and up and down for column flight.

\begin{figure}
\centering
\includegraphics[width=\columnwidth]{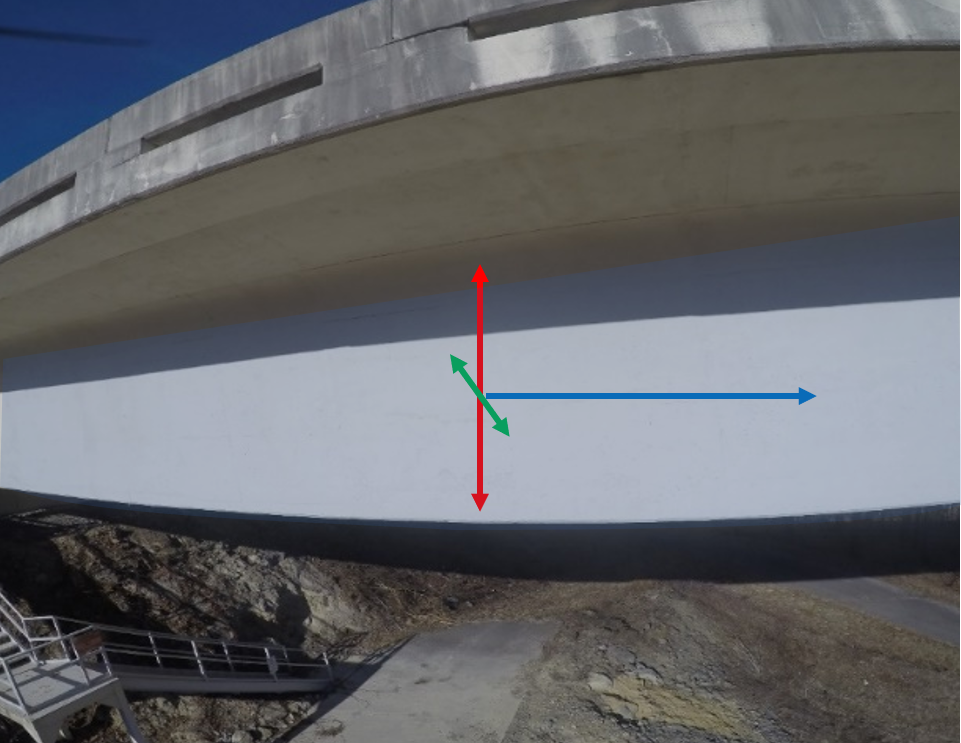}
\caption{Independent PID loops maintain position along the red and green axes, while a constant velocity is given along the blue axis.}
\label{controlc}
\end{figure}

As an example, for the flight \textit{GR} (Fig.~\ref{controlc}):
\begin{itemize}
\item One PID loop maintains the desired distance along the axis perpendicular to the girder surface (green axis). 
\item The second PID loop maintains altitude w.r.t the top of the bridge girder (red axis).
\item A nominal velocity drives the UAV along the axis parallel to the girder surface (blue axis).
\end{itemize}

\subsubsection{Supervisor}\label{sec:supervisor}
The role of the supervisor is to make the decision of when to switch between local navigation routines. The supervisor executes autonomous switching between local navigation routines in the order that is given by the planner, described in Section~\ref{sec:planner}.

\begin{figure}[hbt!]
\centering
\includegraphics[width=0.99\columnwidth]{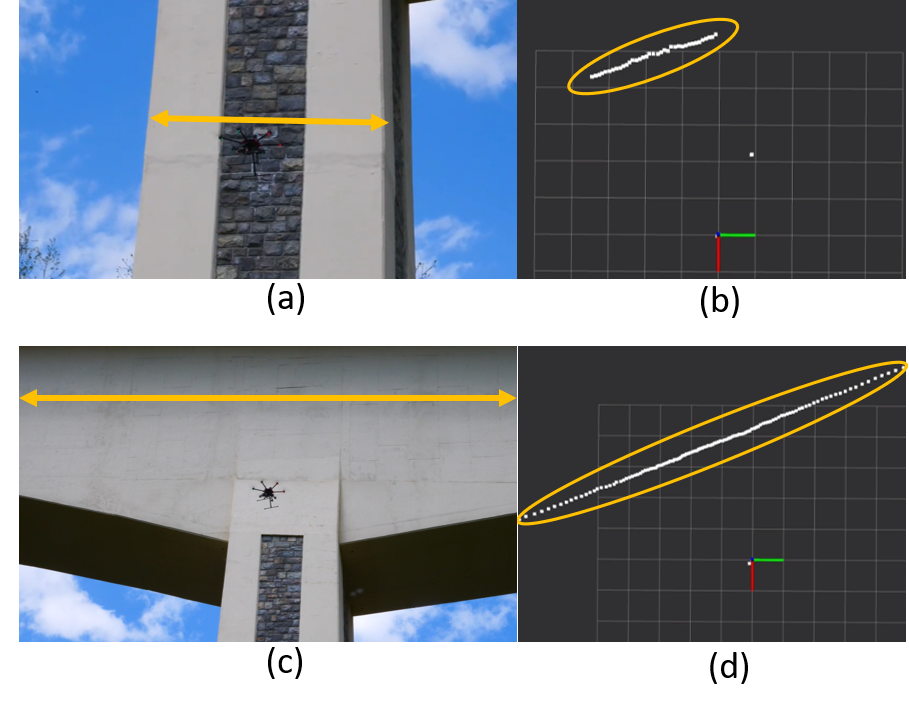}
\caption{Switching from column follow to girder follow.}
\label{c2g}
\end{figure}

When making the decision to switch between local navigation routines we look for special characteristics in the data from the two 2D \textsc{Lidar}s. This allows us to identify when to switch between local navigation routines. For instance, consider the scenario where the UAV is traversing up a bridge column using the local navigation routine \textit{CU}, as shown in Figs.~\ref{c2g}(a)(b). When analyzing this behavior we use the horizontal \textsc{Lidar} data to determine if the UAV can switch routines. When executing \textit{CU} along the column of the bridge the horizontal \textsc{Lidar} data is constant. However when encountering the bridge girder, the number of data points from the horizontal \textsc{Lidar} increases dramatically as shown in Figs.~\ref{c2g}(c)(d). Due to this behavior, the supervisor knows that the UAV can autonomously switch from \textit{CU} routine to the either \textit{GR} or \textit{GL} routines.

A coarse estimation of the global position can be used to better inform the supervisor. If we know the global position of the UAV, we can use that information to decide when to switch. However, due to the inaccuracy of GPS around bridge structures, we remove all GPS data for our simulations and experiments.

It is to be noted that the supervisor need not necessarily be fully automated even though we implemented a fully autonomous method. Currently the supervisor executes fully autonomous flight for 2D surfaces at a time, for instance one surface of the bridge containing multiple sections of girders and columns. Once the UAV has completed coverage of one bridge surface using the local navigation routines, a human operator would then switch the UAV to another bridge surface. This can be done if needed and in tricky scenarios such as switching from flight under the bridge to flight beside the bridge. This allows execution of 3D coverage of the bridge autonomously with a minimum human operator intervention.

\section{Experiments and Results}\label{exp}

\subsection{Individual Routines}
\begin{figure}[hbt!]
\centering
\includegraphics[width=\columnwidth]{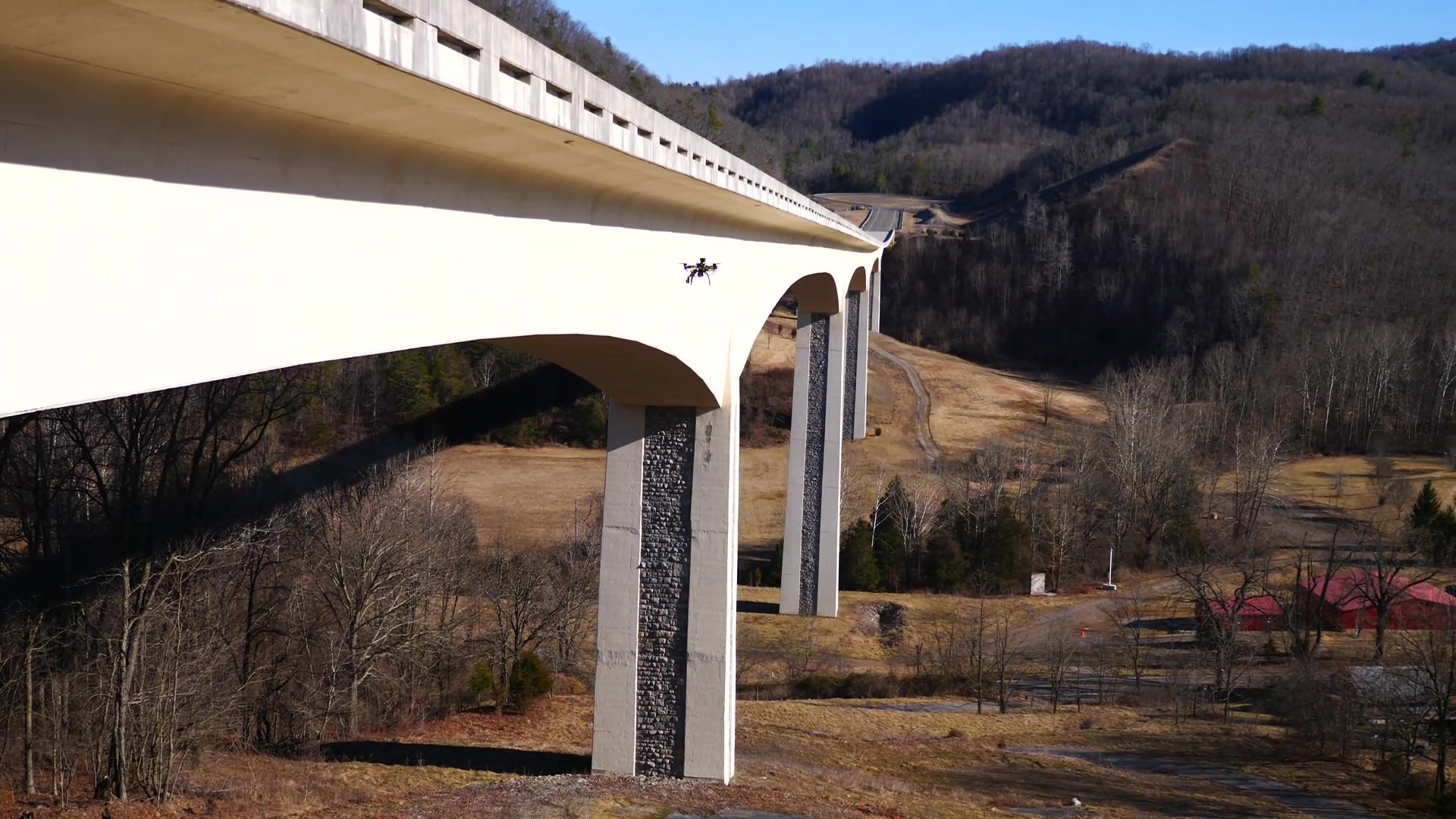}
\caption{Experiment at the VTTI Smart Road Bridge.}
\label{vtti_exp}
\end{figure}

We conducted experimental flights at the Virginia Tech Transportation Institute (VTTI) Smart Road Bridge, which is 53m tall, 609m long, 5 span, variable height concrete box girder bridge (Fig.~\ref{vtti_exp}). We tested \textsc{Lidar} based autonomous flight (no GPS) with a GoPro camera onboard collecting images of the bridge.

\subsubsection{Flight beside bridge girder}
The objective of the experiment was to have the UAV fly alongside the bridge (i.e., \textit{GR}) with a nominal velocity of 0.5m/s maintaining a horizontal separation of 4.5m to the girder. The \textsc{Lidar} scan (rotation) rate was 2Hz, and hence the control signal (velocity in the direction perpendicular to the bridge) was also issued at 2Hz. The \textsc{Lidar} can scan at up to 10Hz, and a higher control rate can be employed, for instance, in windy conditions. The algorithmic computations only take approximately 1ms. Fig.~\ref{beside_hor} shows the control signal driving the UAV to the desired separation of 4.5m from the structure. Hence, we could experimentally verify that accurate flight alongside the bridge at a fixed horizontal distance is feasible, which is helpful for consistent data (image) collection during flight.

Maintaining the correct altitude with respect to the bridge structure proved to be more challenging. Since the terrain around the bridge is not even, using a downward facing \textsc{Lidar} is not an option for maintaining altitude. We cannot rely on the barometer alone, not only because of barometric drifts and inaccuracies, but also because the bridge may slope up or down. Hence, we need some method to hold position w.r.t. the bridge structure itself. For this purpose, we use the vertical cut from the 2D \textsc{Lidar} rotating in the vertical plane that gives us a cross-section of the bridge. The vertical cut was used to maintain an altitude of 2.5m below the top of the girder as shown in Fig.~\ref{beside_vert}. 

\subsubsection{Flight along bridge column}
\begin{figure}[hbt!]
\centering
\begin{subfigure}[b]{0.45\textwidth}
\centering
\includegraphics[width=\columnwidth]{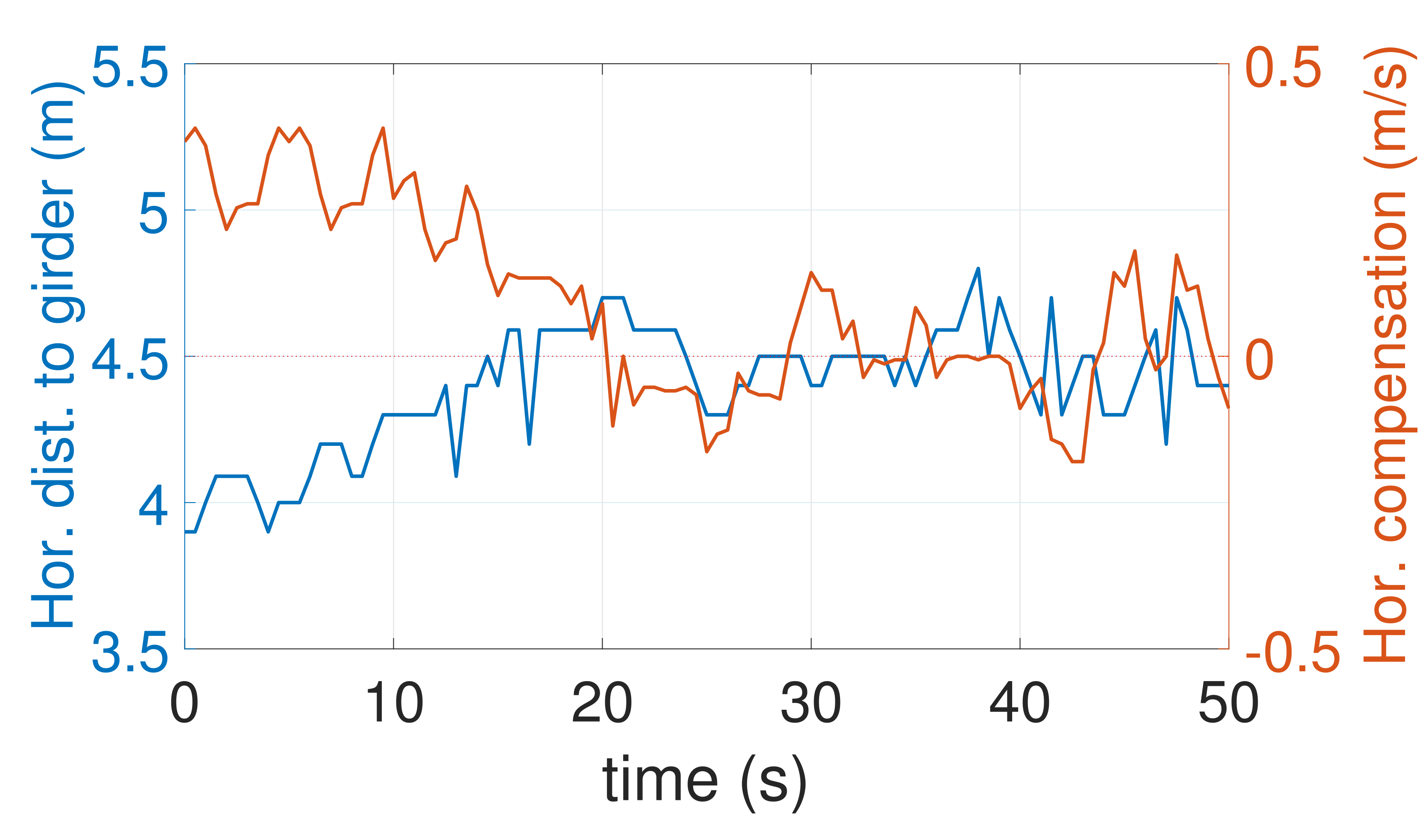}
\caption{}
\label{beside_hor}
\end{subfigure}
\hfill
\begin{subfigure}[b]{0.45\textwidth}
\centering
\includegraphics[width=\columnwidth]{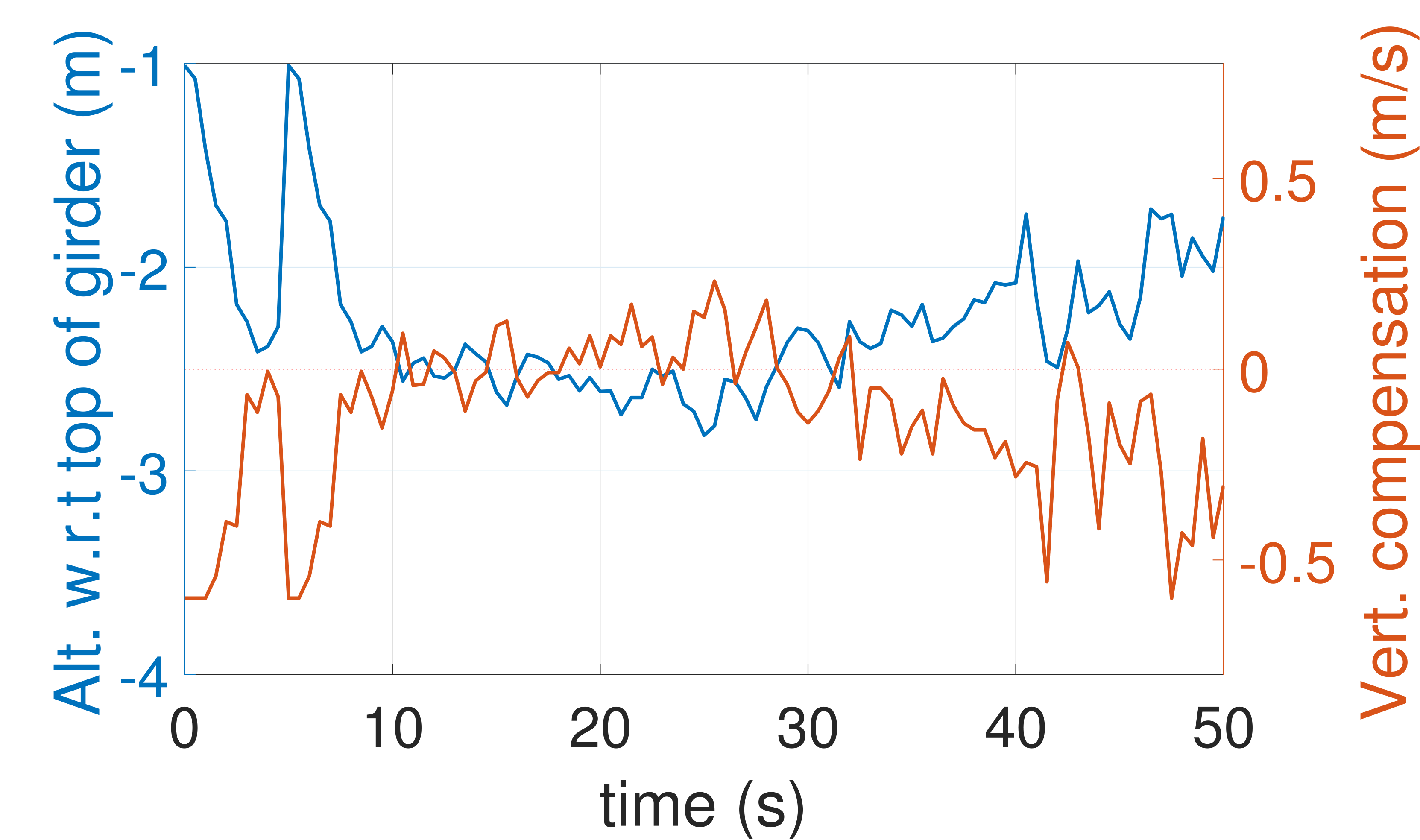}
\caption{}
\label{beside_vert}
\end{subfigure}
\hfill
\begin{subfigure}[b]{0.45\textwidth}
\centering
\includegraphics[width=\columnwidth]{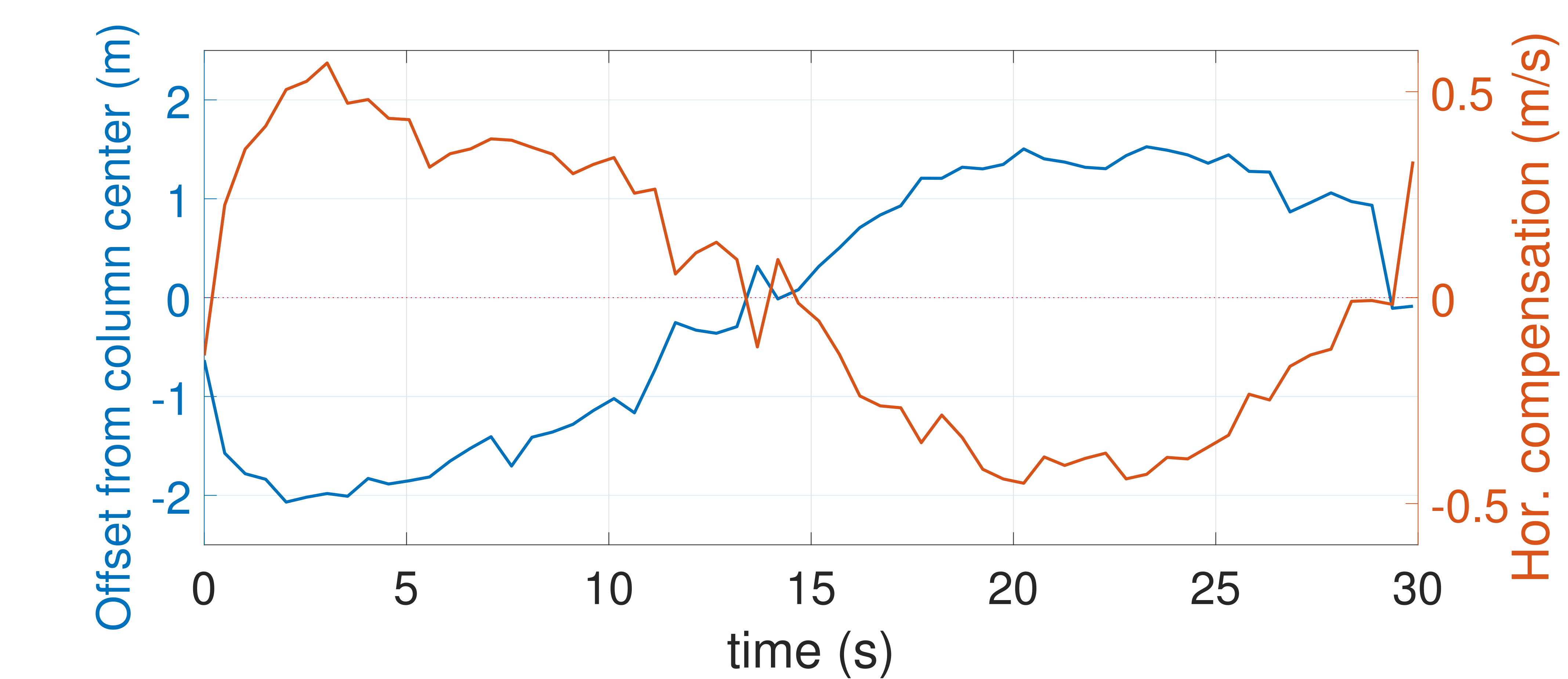}
\caption{}
\label{column_1}
\end{subfigure}
\hfill
\begin{subfigure}[b]{0.45\textwidth}
\centering
\includegraphics[width=\columnwidth]{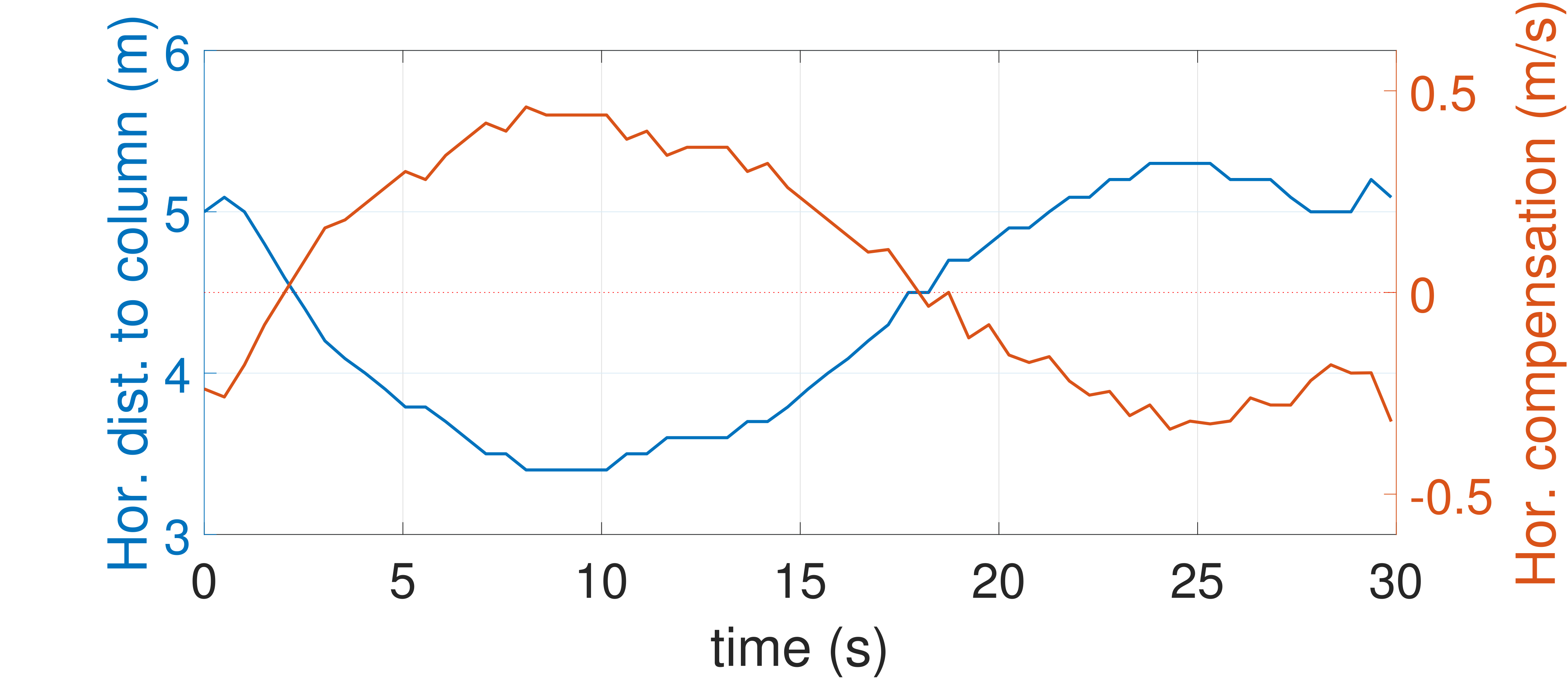}
\caption{}
\label{column_2}
\end{subfigure}
\caption{(a) Correction velocity in the direction perpendicular to the bridge structure maintaining the desired distance of 4.5m from it. (b) Correction velocity in the vertical direction maintaining the desired altitude of 2.5m below the top of the girder. (c) Correction velocity in the direction tangential to the bridge column centering the UAV w.r.t the column. (d) Correction velocity in the direction perpendicular to the bridge structure maintaining the desired distance of 4.5m from it.}
\end{figure}

The objective of this experiment was to have the UAV execute \textit{CU} and \textit{CD} along the bridge column with a vertical nominal velocity of 0.5m/s or -0.5m/s,  while compensating in the horizontal direction. We maintain a separation of 4.5m w.r.t. the column. The experimental results in Figs.~\ref{column_1} \&~\ref{column_2} show the control algorithm compensating to maintain the center and the desired distance to the column. 

\subsection{Full Scale Sims}
\begin{figure}[hbt!]
    \centering
    \includegraphics[width=\columnwidth]{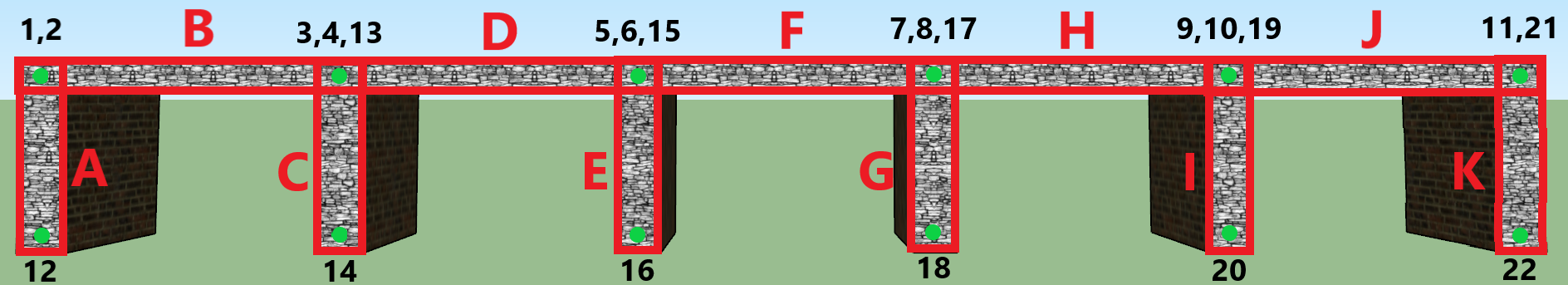}
    \caption{Bridge used for full scale simulations split into planner surfaces. The red letters represent the polygon surfaces and black numbers are the ID of the green nodes representing the coverage of each planner surfaces. \textit{GL} and \textit{GR} local navigation routines can be executed on polygon surfaces B, D, F, H, and J. \textit{CU} and \textit{CD} local navigation routines can be executed on polygon surfaces A, C, E, G, I, and K.}
    \label{fig:fullBridgePlanner}
\end{figure}

In addition to conducting experiments for the individual routines, we conduct simulation experiments on full scale bridges. These bridges are modeled in Gazebo 7. We start by creating a bridge in SketchUp and exporting a .dae file to create a model in Gazebo. In Fig.~\ref{fig:fullBridgePlanner} we show the full image of our bridge with annotations to represent individual sections. In Fig.~\ref{fig:fullBridgePlanner} the green nodes represent possible locations of switching and the red outlines represent the polygons created. We obtain the optimal tour output from our GTSP planner. As a reminder we initially feed the planner a set of planar surfaces, shown in Fig.~\ref{fig:fullBridgePlanner}. We use Fig.~\ref{fig:fullBridgePlanner} to obtain a rough estimate of the distances between entrance and exit nodes for each planner surface. This means that the edge costs are not the actual bridge distances, but a scaled version. The reason for a non-exact measurement is to replicate what a human inspector/operator might have on hand at a work site. Since inspectors do not always have access to exact specifications of a bridge they might have to take an overview picture and then make rough estimates on the relationships between entrance and exit nodes in the picture.

\begin{figure}[hbt!]
    \centering
    \includegraphics[trim = 10 1 10 10, clip, width=\columnwidth]{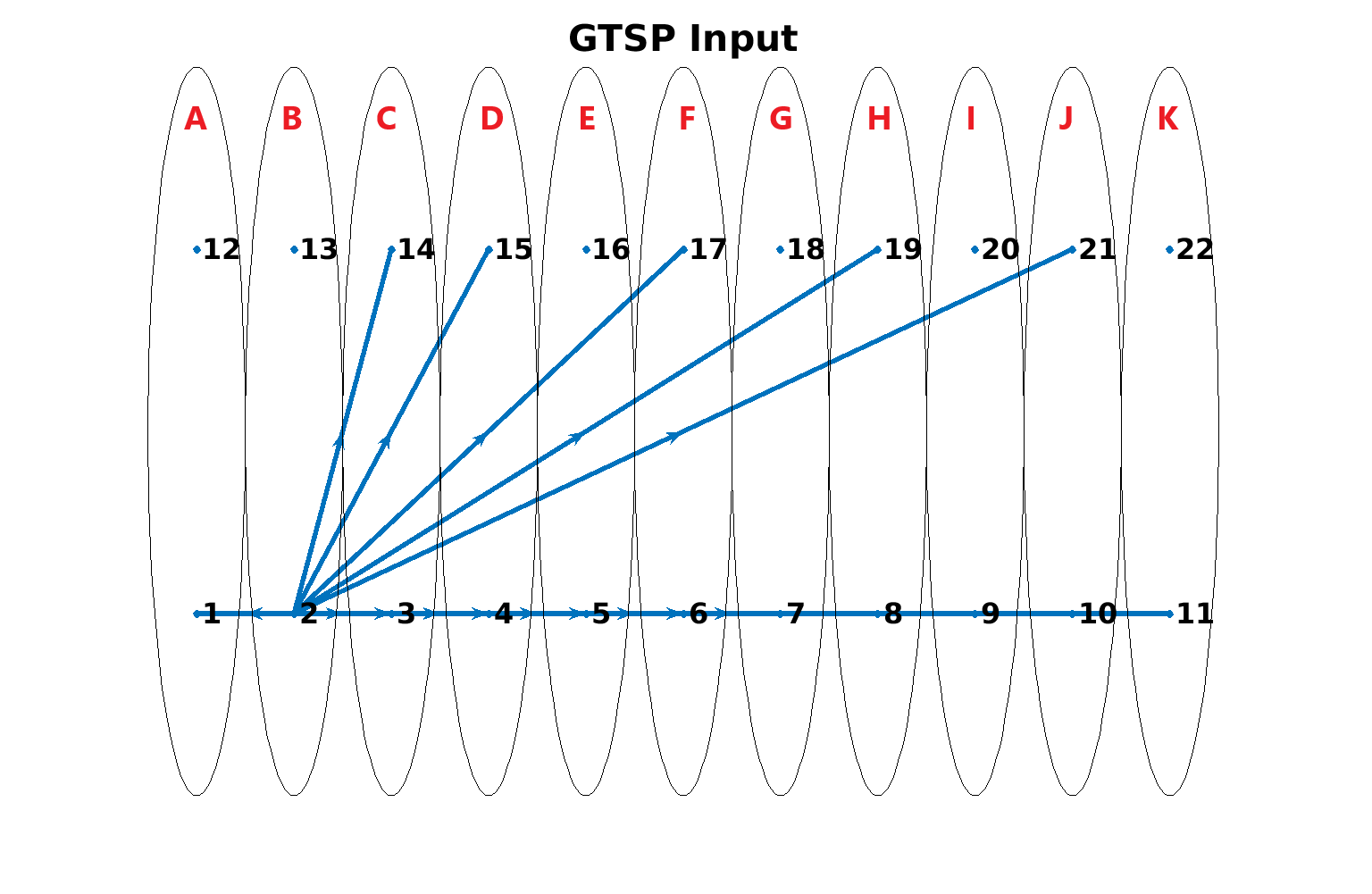}
    \caption{Input to GTSP solver. The red letters correspond to the polygon identification numbers in Fig.~\ref{fig:fullBridgePlanner} and the black numbers correspond to the green nodes created for each polygon to signify the direction of coverage in Fig.~\ref{fig:fullBridgePlanner}. Here is a subset of the edges that are created as input to the GTSP solver. We do not display all edges for readability.}
    \label{fig:gtspInput}
\end{figure}

\begin{figure}[hbt!]
    \centering
    \includegraphics[trim = 10 1 10 10, clip, width=\columnwidth]{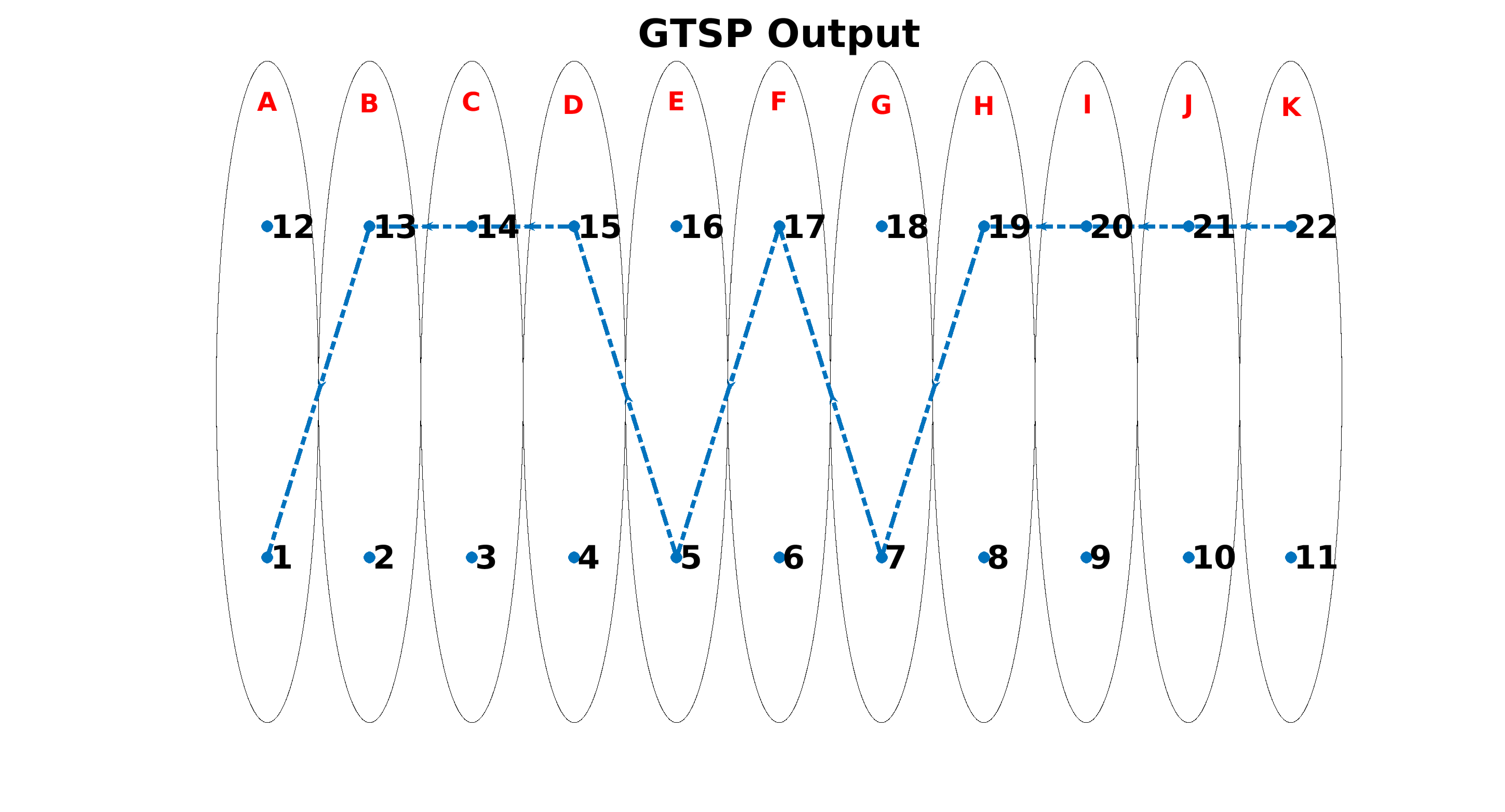}
    \caption{Output from GTSP solver. The red letters correspond to the polygon identification numbers in Fig.~\ref{fig:fullBridgePlanner} and the black numbers correspond to the green nodes created for each polygon to signify the direction of coverage in Fig.~\ref{fig:fullBridgePlanner}. The solver starts with polygon K and moves from right to left. With this the output of the GTSP solver gives \textit{CU} , \textit{GL}, \textit{CD} and repeats until the the last polygon A ending on \textit{CD}.}
    \label{fig:gtspOutput}
\end{figure}

We start by converting the instance shown in Fig.~\ref{fig:fullBridgePlanner} into a GTSP instance, shown in Fig.~\ref{fig:gtspInput}. In Fig.~\ref{fig:gtspInput} we show the clusters created as well as a subset of edges created and the costs. When creating the GTSP instance a fully connected graph is created and then edges are pruned out. The edges that are removed are inter-cluster edges and edges that are deemed impossible because the UAV would lose reference to the bridge structure, not being able to navigate correctly. Once this GTSP instance is created the input can be given to a GTSP solver, GLNS, and an output is obtained.

Once we have obtained the output of the order in which we should visit from the planner, shown in Fig.~\ref{fig:gtspOutput}, we give it to the supervisor. Fig.~\ref{fig:gtspOutput} shows the output from the planner. Each cluster is shown with the ID letter from Fig.~\ref{fig:gtspOutput} matching the ones in Fig.~\ref{fig:fullBridgePlanner} as well as the node ID numbers. Once the supervisor has the output from the planner it is able to execute the full coverage of the bridge's planner surface. For the bridge in Fig.~\ref{fig:fullBridgePlanner} the planner took 1.92s and gave us the solution of \textit{CU} on column K, \textit{GL} on girder J, \textit{CD} on column I, and then repeating this until column A is reached. This order of local navigation routines is given to the supervisor and executed in real time.

\begin{figure}[hbt!]
    \centering
    \includegraphics[width=\columnwidth]{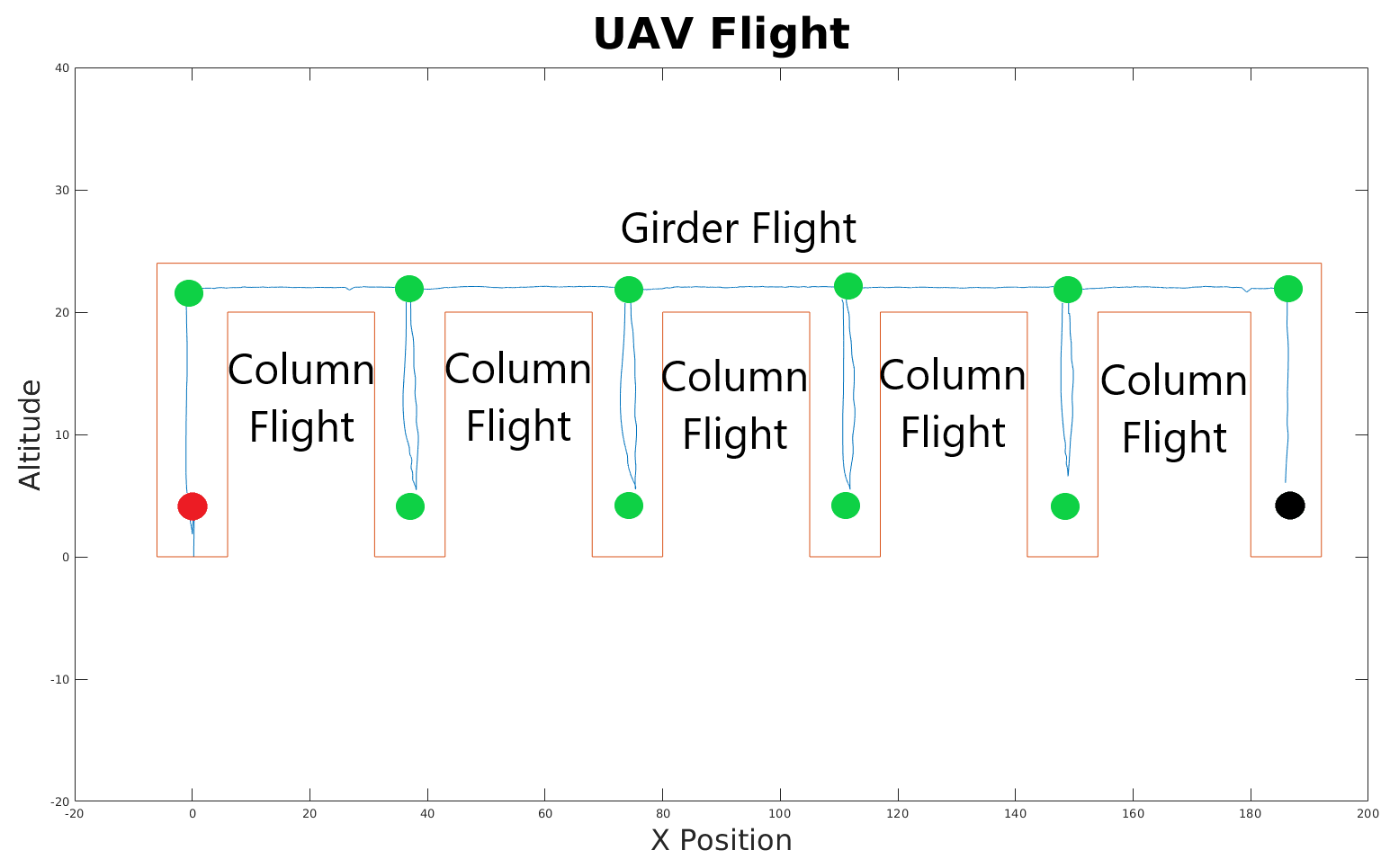}
    \caption{Simulation data obtained from mavros of UAV flight on the bridge located in Fig.~\ref{fig:fullBridgePlanner}. The blue line indicates the actual flight of the UAV and the orange lines represent the bridge structure. The green nodes are locations when the UAV switched modes from girder to column flight or \textit{CD} to \textit{CU} flight. The black and red nodes represent the starting and end location of the UAV path, respectively.}
    \label{fig:bridge5Output}
\end{figure}

We can see in Fig.~\ref{fig:bridge5Output} the path taken by the UAV. We plot the GPS location of the UAV from the simulation in Matlab and annotate the figure. In the figure the green nodes are where the UAV Scheduler switched between local navigation routines. The black and red nodes represent the start and end of the UAVs path for coverage of the bridge. This proof-of-concept implementation demonstrates the feasibility of the entire system.

\section{Conclusion and Future Work} \label{future}
We presented a fast  LIDAR based approach for autonomous navigation using two 2D \textsc{Lidars} scanning in horizontal and vertical planes. We described strategies to switch between navigation routines to cover various bridge surfaces. We envision full coverage of the bridge using a total of eight routines for UAV flight. \textit{GR}, \textit{GL}, \textit{CU}, and \textit{CD} have already been tested and the rest, characterized in Section~\ref{lidar}, are conceptually similar to the ones already tested. The \textsc{Lidar} based approach can also be used to provide safety guarantees in assisted manual flights. Operator (pilot) input can be rejected if it will result in the UAV getting too close to the bridge structure. Adapting ideas from our present approach to different types of bridges (which could require working with other sensors such as Stereo camera for navigation) is a future direction.

\section{acknowledgement}
This work has been funded in part by the Center for Unmanned Aircraft Systems (C-UAS), a National Science Foundation-sponsored industry/university cooperative research center (I/UCRC) under NSF Award No. IIP-1161036 along with significant contributions from C-UAS industry members as well as NSF Award No. 1840044.

\bibliographystyle{unsrt}
\bibliography{main.bib}
\end{document}